\documentclass{sig-alternate-05-2015}
\usepackage{breqn}
\usepackage{hyperref}
\usepackage{empheq}
\usepackage{multicol,lipsum}

\usepackage{amsmath}
\usepackage{txfonts}
\usepackage{subfigure} 
\usepackage{multirow}
\usepackage{hhline}
\usepackage{multirow}
\usepackage{rotating}
\usepackage[utf8]{inputenc}
\usepackage{color}

\begin{document}

\setcopyright{acmcopyright}


\doi{http://dx.doi.org/10.1145/3056540.3064942}
    
\isbn{978-1-4503-5227-7/17/06}
      
\conferenceinfo{PETRA '17}{June 21-23, 2017,Island of Rhodes,Greece}

\acmPrice{\$15.00}

%

\title{Improving the Accuracy of the CogniLearn System for Cognitive Behavior Assessment}
%
%
%
%
%

\numberofauthors{6} 
%
\author{
%
%
\alignauthor
Amir Ghaderi\\
       \affaddr{University of Texas at Arlington}\\
       \affaddr{Texas, USA}\\
       \email{amir.ghaderi@mavs.uta.edu}
\alignauthor
Srujana Gattupalli
       \affaddr{University of Texas at Arlington}\\
       \affaddr{Texas, USA}\\
       \email{srujana.gattupalli@\\mavs.uta.edu}
\alignauthor Dylan Ebert \\
       \affaddr{University of Texas at Arlington}\\
       \affaddr{Texas, USA}\\
       \email{dylan.ebert@mavs.uta.edu}
\and  
\alignauthor Ali Sharifara\\
        \affaddr{University of Texas at Arlington}\\
       \affaddr{Texas, USA}\\
       \email{ali.sharifara@uta.edu}
\alignauthor Vassilis Athitsos\\
       \affaddr{University of Texas at Arlington}\\
       \affaddr{Texas, USA}\\
       \email{athitsos@uta.edu}
\alignauthor Fillia Makedon\\
      \affaddr{University of Texas at Arlington}\\
       \affaddr{Texas, USA}\\
       \email{makedon@uta.edu}
}

\maketitle
\begin{abstract}
HTKS \cite{mcclelland2014predictors} is a game-like cognitive assessment method, designed for children between four and eight years of age. During the HTKS assessment, a child responds to a sequence of requests, such as ``touch your head'' or ``touch your toes''. The cognitive challenge stems from the fact that the children are instructed to interpret these requests not literally, but by touching a different body part than the one stated. In prior work, we have developed the CogniLearn system, that captures data from subjects performing the HTKS game, and analyzes the motion of the subjects. In this paper we propose some specific improvements that make the motion analysis module more accurate. As a result of these improvements, the accuracy in recognizing cases where subjects touch their toes has gone from 76.46\% in our previous work to 97.19\% in this paper.
\end{abstract}

%
%


%
%

%
%



\section{Introduction}
HTKS \cite{mcclelland2012self,mcclelland2014predictors} is a game-like cognitive assessment method, designed for children between four and eight years of age. During the HTKS assessment, a child responds to a sequence of requests, such as ``touch your head'' or ``touch your toes''. The cognitive challenge stems from the fact that the children are instructed to interpret these requests not literally, but by touching a different body part than the one stated. For example, a child may be instructed to touch her toes in response to the request ``touch your head''. HTKS has been shown to be related to measures of cognitive flexibility, working memory, and inhibitory control. At the same time, HTKS has also been shown to be useful in predicting academic achievement growth for prekindergarten and kindergarten children.

In our prior work, we have developed the CogniLearn system \cite{cognilearn}, that can be used to record the motion of human subjects as they play the HTKS game, and that also analyzes that motion so as to assess how accurately the subjects executed the requested tasks. In CogniLearn, a \textit{Microsoft Kinect V2} camera is used for recording human motion. Then, we use the DeeperCut method \cite{insafutdinov2016deepercut} to perform body pose estimation in each frame. Finally, using the body pose estimates from DeeperCut we use a classification module that determines whether the subject touched his or her head, shoulders, knees, or toes. The CogniLearn system compares the part that was touched with the part that should have been touched based on the rules of the game, and assesses the overall accuracy score of the person playing the game.

The rest of the paper is organized as follows: In Section \ref{section_related} we discuss related work in this area. In Section \ref{section_method} we describe the proposed improvements to the prior version of the CogniLearn system. A quantitative evaluation of these improvements is offered in the experiments (Section \ref{section_experiments}).

\section{Related Work}
\label{section_related}

Several deep-learning methods have been proposed in recent years for video analysis and activity recognition \cite{donahue2015long,gattupalli2016evaluation,scalable}, offering significantly improved accuracy compared to previous approaches\cite{leightley2013human,xia2013spatio}. Deep learning methods have also been used in supervised or unsupervised manner in different tasks in computer vision \cite{insafutdinov2016deepercut, ghaderi2016selective}, oftentimes producing state-of-the-art results.

In \cite{cognilearn} we have introduced the CogniLearn system, which is used for automated video capture and  performance assessment during the HTKS assessment. CogniLearn is designed to provide meaningful data and  measures that can benefit therapists and cognitive experts. More specifically, the motion analysis and evaluation module provides systematic feedback regarding the performance of the HTKS tasks to the human experts. In this paper, we build upon the CogniLearn system, and we suggest some specific improvements in the motion analysis module, that lead to higher recognition accuracy.
\\
\begin{figure}
\centering
\includegraphics[height=2in, width=2in]{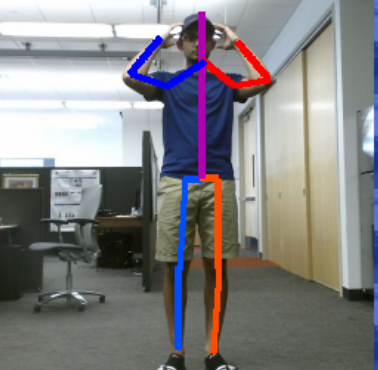}
\caption{A sample human body pose estimation on a frame using DeeperCut \cite{insafutdinov2016deepercut}.}
\label{fig:humanBodyEstimatedFrame}
\end{figure}

\section{Our Method}
\label{section_method}
We use DeeperCut \cite{insafutdinov2016deepercut} to estimate the location of human body parts in each color frame of the video. Figure \ref{fig:humanBodyEstimatedFrame} shows a video frame where we have superimposed the body part locations estimated by DeeperCut. Each color frame of a test video sequence is provided as input to the DeeperCut method. The output of the algorithm is the image location of 12 body parts: head, shoulder(right and left), elbow(right and left), wrist(right and left), hip, knee(right and left), ankle(right and left).

After we obtain the body part locations from DeeperCut, we perform an additional step, in order to estimate whether the human, at that frame, is touching his or her head, shoulders, knees, or toes. As a first step, we define a distance $D$ between hands and head, hands and shoulder, hands and knees, and hands and ankles. Using $\|\cdot\|$ to denote Euclidean norms, this distance is defined as follows:

\begin{equation}
D(\mathrm{head})= \frac{\|\mathrm{lh} - \mathrm{head}\| + \|\mathrm{rh} - \mathrm{head}\|}{2}
\end{equation}

\begin{equation}
D(\mathrm{shoulders})= \frac{\|\mathrm{lh} - \mathrm{ls}\| + \|\mathrm{rh} - \mathrm{rs}\|}{2}
\end{equation}

\begin{equation}
D(\mathrm{knees})= \frac{\|\mathrm{lh} - \mathrm{lk}\| + \|\mathrm{rh} - \mathrm{rk}\|}{2}
\end{equation}

\begin{equation}
D(\mathrm{ankles})= \frac{\|\mathrm{lh} - \mathrm{la}\| + \|\mathrm{rh} - \mathrm{ra}\|}{2}
\end{equation}

In the above definitions, $\mathrm{head}$ stands for the $(x,y)$ pixel location of the center of the head in the color frame, as estimated by DeeperCut. Similarly, $\mathrm{lh}$ and $\mathrm{rh}$ stand for the locations the left and right hand, $\mathrm{ls}$ and $\mathrm{rs}$ stand for the locations of the left and right shoulder, $\mathrm{lk}$ and $\mathrm{rk}$ stand for the locations of the left and right knee, and $\mathrm{la}$ and $\mathrm{ra}$ stand for the locations of the left and right ankle.For example, $\|\mathrm{lh} - \mathrm{head}\|$ denotes the Euclidean distance between the left hand and the center of the head.

Based on these $D$ values, one approach for estimating the body part that is being touched is to simply select the body part for which the $D$ score is the smallest. This was the approach used in \cite{cognilearn}. However, when the person touches the toes or knees, this approach does not work well. When a person bends down to touch the knees or toes with the hands, the head inevitably also gets near to the knees or toes. In that case, two issues may arise. The first one is that the accuracy of the body joint estimator is decreased. The second issue is that the  detected location for the head is near the detected locations for the knees or toes. As a result, for example, when the hands are touching the toes, it frequently happens that the distance of hands to the head is estimated to be smaller than distance of the hands to the toes. These two issues can lead to inaccuracies. As we see in Table \ref{tab:oldRes}, in the original CogniLearn results of \cite{cognilearn}, 9.33\% of toe frames are classified as head frames, and 14.00\% of toe frames are classified as knee frames.

In this paper, we propose two heuristic rules to improve the classification accuracy of toe frames: \\

\textbf{Rule 1}: If the distance between the head and the hip is less than a predefined threshold, we can immediately conclude that the hands are touching the toes. \\ 

\textbf{Rule 2}: Sometimes, when the hands are touching the head, the distance between the hands and the head is estimated to be longer than the distance between the hand and the shoulders. To address this issue, we add a constant bias value to the distance between hands and shoulders, before comparing it with the distance between the hands and the head.
\\

In the experiments, we demonstrate that these two rules significantly improve the classification accuracy on toe and head frames, while only minimally affecting the classification accuracy on frames where the hands touch the shoulders or knees.

\section{Experiments}
\label{section_experiments}
\begin{figure}[t] 
\centering
  \subfigure[ ]{%
    \includegraphics[width=0.16\textwidth]{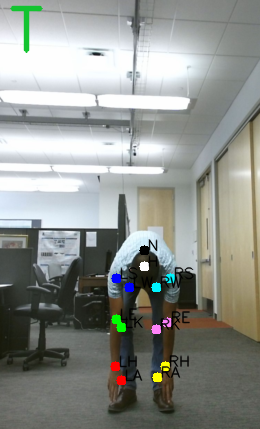} 
  } 
  \quad
  \subfigure[ ]{%
    \includegraphics[width=.16\textwidth]{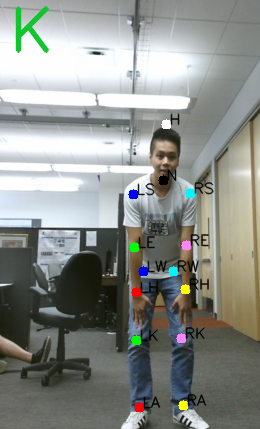} 
  }
  \caption{Results using the full method described in this paper, i.e., when both Rule 1 and Rule 2 are used. On the left, we see a frame where the hands touch the toes. On the right, we see a frame where the hands touch the knees. The green letter on the top left of each frame is the classification output of the system, where ``T'' stands for ``toes'', ``K'' stands for ``knees''.}
\label{fig:example}
\end{figure}

For our experiments, we use the same dataset that was used in the original CogniLearn paper  \cite{cognilearn}. The dataset includes color videos from 15 participants, whose ages are between 18 and 30 years (while the HTKS assessment has been designed for children between the ages of 4 and 8, at this time we still do not have recorded data available from children of that age). In total, the dataset contains over 60,000 video frames. Figure \ref{fig:example} shows examples of test frames correctly recognized by our algorithm. The green letter in top left of the images shows the classification output of our system (``T'' stands for ``toes'', ``K'' stands for ``knees''). 

Our method is applied on each color frame separately.
The goal of our method is to classify each frame into one of four classes, corresponding respectively to whether the human is touching his or her head, shoulders, knees, or toes. Ground truth information is provided for 4,443 video frames, and we use those frames as our test set. The ground truth  specifies, for each frame, which of the four classes belongs to. Accuracy is simply measured as the percentage of test frames for which the output of our system matched the ground truth.

We should emphasize that the results that we present are user-independent. None of the 15 subjects appearing in the test set is used to train any part of our models. The only module that uses training is DeeperCut, and we use the pretrained model that has been made available by the authors of \cite{insafutdinov2016deepercut}.

\subsection{Results}


Table \ref{tab:oldRes} shows the confusion matrix reported in the original CogniLearn paper \cite{cognilearn}. As we can see in that table, shoulder and knee frames are recognized at rather high accuracies of 99.63\% and  98.17\% respectively. However, head and toes frames are recognized with lower accuracies, 94.47\% and 76.46\% respectively. This paper was primarily motivated by the need to improve the accuracy for those two cases.

\begin{table}[!th] 

\caption{Confusion matrix reported by \cite{cognilearn}. Rows correspond to ground truth labels, and columns correspond to classification outputs.} 
\label{tab:oldRes}
  \centering
  \begin{tabular}{|c|c|c|c|c|c|c|}
    \cline{3-6} 
    \multicolumn{2}{c|}{} & \multicolumn{4}{c|}{Recognized} & \multicolumn{1}{|c}{} \\ 
    \cline{3-7} 
    \multicolumn{2}{c|}{} &  Head & Shoulder & Knee & Toe & \footnotesize{Sum} \\ \hline

    \multirow{4}{*}{\begin{turn}{-90} Real \end{turn}} & Head & \textbf{94.47} &5.53 & 0.00& 0.00 & \small{100}\\  \cline{2-7}  & Shoulder  & 0.12& \textbf{99.63} &0.25& 0.00 & \small{100} \\ \cline{2-7}
           & Knee  & 0.00 & 0.54 & \textbf{98.17} & 1.29 & \small{100}  \\ \cline{2-7}
           & Toe   & 9.33 & 0.21 & 14.00 & \textbf{76.46} & \small{100}  \\ \cline{2-7}
           \hline
  \end{tabular}
\end{table} 
In Table \ref{table:ruleOneTwoRes} we report the results from the method proposed in this paper (i.e, when we apply both Rule 1 and Rule 2 from Section \ref{section_method}. As we can see, the accuracy for all four categories is more than 94.7\%. The accuracy for head frames is marginally improved compared to \cite{cognilearn}. The accuracy for shoulder and knee frames is slightly worse compared to \cite{cognilearn}. At the same time, the accuracy for toe frames is now 97.19\%, significantly higher than the accuracy of 76.46\% reported in \cite{cognilearn}.
\begin{table}[!t] 
\caption{Confusion matrix obtained using the full method described in this paper, i.e., when both Rule 1 and Rule 2 are added to the method of \cite{cognilearn}. Rows correspond to ground truth labels, and columns correspond to classification outputs.}
\label{table:ruleOneTwoRes}
  \centering
  \begin{tabular}{|c|c|c|c|c|c|c|}
    \cline{3-6} 
    \multicolumn{2}{c|}{} & \multicolumn{4}{c|}{Recognized} & \multicolumn{1}{|c}{} \\ 
    \cline{3-7} 
    \multicolumn{2}{c|}{} &  Head & Shoulder & Knee & Toe & \footnotesize{Sum} \\ \hline

    \multirow{4}{*}{\begin{turn}{-90} Real \end{turn}} & Head & \textbf{94.78} & 3.39 & 0.26& 1.57 & \small{100}\\  \cline{2-7}  & Shoulder  & 0.50& \textbf{99.25} &0.12& 0.12 & \small{100} \\ \cline{2-7}
           & Knee  & 0.00 & 0.60 & \textbf{97.22} & 2.18 & \small{100}  \\ \cline{2-7}
           & Toe   & 0.76 & 0.00 & 2.05 & \textbf{97.19} & \small{100}  \\ \cline{2-7}
           \hline
  \end{tabular}
\end{table} 
Finally, in Table \ref{tab:ruleOneRes} we show results using a partial implementation of our method, applying only Rule 1, and not Rule 2. We note that the overall accuracy is mostly similar to what we get when we combine Rules 1 and 2. Overall, Rule 1 is by far the biggest contributor to the improvements we obtain over the original results of \cite{cognilearn}. At the same time, the accuracy for head frames improves from 93.21\% to 94.78\% when we use Rules 1 and 2, compared to using only Rule 1. Rule 2 was explicitly designed to reduce the percentage of head frames that were classified as shoulder frames. Indeed, using Rule 2 (together with Rule 1) reduces that percentage from 4.96\% (obtained using only Rule 1) to 3.39\%.
\begin{table}[!t]
\caption{Confusion matrix obtained by adding Rule 1 to the method of \cite{cognilearn}. Rows correspond to ground truth labels, and columns correspond to classification outputs.}
\label{tab:ruleOneRes}
  \centering
  \begin{tabular}{|c|c|c|c|c|c|c|}
    \cline{3-6} 
    \multicolumn{2}{c|}{} & \multicolumn{4}{c|}{Recognized} & \multicolumn{1}{|c}{} \\ 
    \cline{3-7} 
    \multicolumn{2}{c|}{} &  Head & Shoulder & Knee & Toe & \footnotesize{Sum} \\ \hline

    \multirow{4}{*}{\begin{turn}{-90} Real \end{turn}} & Head & \textbf{93.21} & 4.96 & 0.26 & 1.57 & \small{100}\\  \cline{2-7}  & Shoulder  & 0.37& \textbf{99.39} &0.12& 0.12 & \small{100} \\ \cline{2-7}
           & Knee  & 0.00 & 0.60 & \textbf{97.22} & 2.18 & \small{100}  \\ \cline{2-7}
           & Toe   & 0.76 & 0.00 & 2.05 & \textbf{97.19} & \small{100}  \\ \cline{2-7}
           \hline
  \end{tabular}
\end{table} 
Table \ref{tab:overall} shows the overall classification accuracy.  In that table, the overall accuracy is defined as the average of the accuracies over the four different classes. The overall accuracy improves from the 92.18\% rate of \cite{cognilearn} to 96.75\% when we add Rule 1, and to 97.11\% when we also add Rule 2.


\begin{table}[ht!]
\caption{Comparisons in accuracy between the original results of \cite{cognilearn}, the results obtained by adding Rule 1 to the method of \cite{cognilearn}, and the results obtained by adding both Rule 1 and Rule 2 to the method of \cite{cognilearn} }
\begin{tabular}{| c | c | c | c | c |c |}  
\hline
  & \textbf{Overall} & \textbf{H} &  \textbf{S} & \textbf{K} & \textbf{T} \\
\hline \hline
Original\cite{cognilearn} & 92.18 & 94.47 & \textbf{99.63} & \textbf{98.17}&76.46 \\ 
\hline
Rule 1 & 96.75 & 93.21 & 99.39 & 97.22 & 97.19  \\ 
\hline
Rules 1,2 combined & \textbf{97.11}  & \textbf{94.78} & 99.25 &  97.22 & \textbf{97.19}  \\ 
\hline
\end{tabular}

\label{tab:overall} 
\end{table}

\begin{figure}[ht] 
\centering
  \subfigure[ ]{%
    \includegraphics[width=0.16\textwidth]{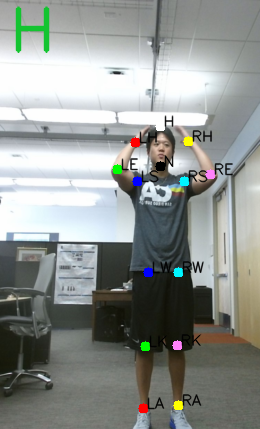} 
  } 
  \quad
  \subfigure[ ]{%
    \includegraphics[width=.16\textwidth]{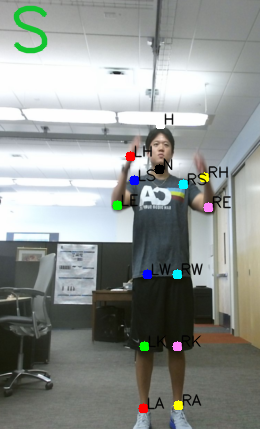} 
  }
  
  \subfigure[]{%
    \includegraphics[width=0.16\textwidth]{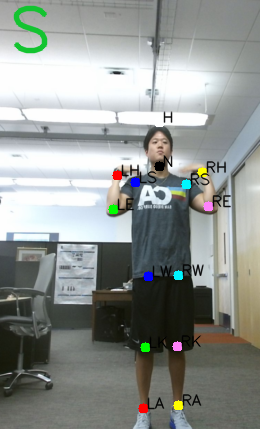} 
  } 
  \quad 
  \subfigure[]{%
    \includegraphics[width=.16\textwidth]{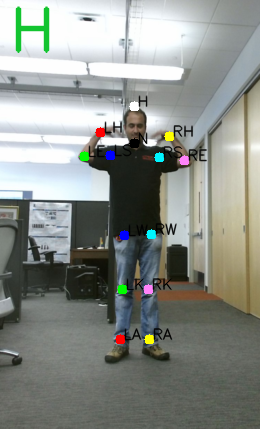} 
  }
  
  \subfigure[]{%
    \includegraphics[width=0.16\textwidth]{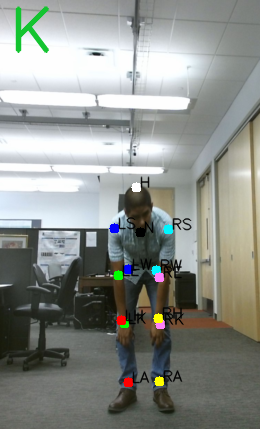} 
  } 
  \quad 
  \subfigure[]{%
    \includegraphics[width=.16\textwidth]{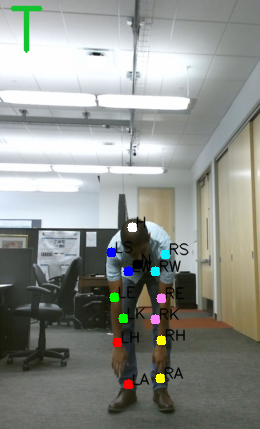} 
  }
  
  \subfigure[]{%
    \includegraphics[width=0.16\textwidth]{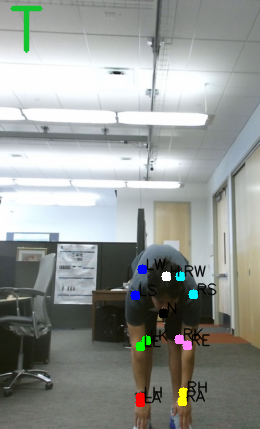} 
  } 
  \quad 
  \subfigure[]{%
    \includegraphics[width=.16\textwidth]{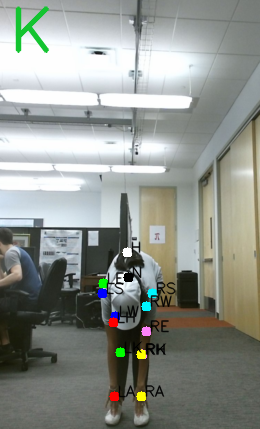} 
  }
  \caption{Example test frames, with the classification output superimposed. The classification output is correct for the examples on the left column, and incorrect for the examples on the right column. The ground truth is: ``head'' for row 1, ``shoulders'' for row 2, ``knees'' for row 3, ``toes'' for row 4.}
\label{samples}
\end{figure}

Figure \ref{samples} shows some sample test frames. More specifically, from each of the four classes we show an example that was classified correctly, and an example that was classified incorrectly. We note that separating the head from the shoulder class can be quite challenging at times, because the distribution of hand positions does not vary much between the two classes. Separating knees and toes can also be difficult, because in frames belonging to both classes the knees are typically occluded, and there is significant overlap between the arms and the legs. This leads to errors in the estimated positions of the hands and the knees.

\section{Conclusions and Future Work}

We have propose a method for improving the accuracy of the original CogniLearn\cite{cognilearn} system in recognizing, for each video frame, whether the human is touching the head, shoulders, knees, or toes in that frame. The experiments have shown that our improvements lead to significantly better accuracy, especially for frames where the human touches the toes. In those cases, the accuracy increased from the 76.46\% rate in \cite{cognilearn} to 97.19\%.

Our project of automatically capturing and analyzing performance in the HTKS test is still in its initial stages. A high priority for us is to obtain data from children between the ages of 4 and 8, as that is the target age group for the HTKS test. Also, we plan to explore using the depth modality of the Kinect camera in addition to the color modality that we have used in \cite{cognilearn} and in this paper. Finally, we should note that the HTKS assessment includes a ``self-correction'' category, in which the subject has started doing an incorrect motion and then self-corrected \cite{mcclelland2014predictors}. In the near future we plan to work on developing methods for identifying such self-correction cases, so that our assessment fully matches the formal HTKS description.


\section*{Acknowledgments}
This work was partially supported by National Science Foundation grants IIS-1055062, CNS-1338118, CNS-1405985, and IIS-1565328. Any opinions, findings, and conclusions or recommendations expressed in this publication are those of the authors, and do not necessarily reflect the views of the National Science Foundation.

%
\bibliographystyle{abbrv}
\bibliography{sigproc}  
%
%
\ifx
\appendix
\section{Headings in Appendices}
The rules about hierarchical headings discussed above for
the body of the article are different in the appendices.
In the \textbf{appendix} environment, the command
\textbf{section} is used to
indicate the start of each Appendix, with alphabetic order
designation (i.e. the first is A, the second B, etc.) and
a title (if you include one).  So, if you need
hierarchical structure
\textit{within} an Appendix, start with \textbf{subsection} as the
highest level. Here is an outline of the body of this
document in Appendix-appropriate form:
\subsection{Introduction}
\subsection{The Body of the Paper}
\subsubsection{Type Changes and  Special Characters}
\subsubsection{Math Equations}
\paragraph{Inline (In-text) Equations}
\paragraph{Display Equations}
\subsubsection{Citations}
\subsubsection{Tables}
\subsubsection{Figures}
\subsubsection{Theorem-like Constructs}
\subsubsection*{A Caveat for the \TeX\ Expert}
\subsection{Conclusions}
\subsection{Acknowledgments}
\subsection{Additional Authors}
This section is inserted by \LaTeX; you do not insert it.
You just add the names and information in the
\texttt{{\char'134}additionalauthors} command at the start
of the document.
\subsection{References}
Generated by bibtex from your ~.bib file.  Run latex,
then bibtex, then latex twice (to resolve references)
to create the ~.bbl file.  Insert that ~.bbl file into
the .tex source file and comment out
the command \texttt{{\char'134}thebibliography}.
\section{More Help for the Hardy}
The sig-alternate.cls file itself is chock-full of succinct
and helpful comments.  If you consider yourself a moderately
experienced to expert user of \LaTeX, you may find reading
it useful but please remember not to change it.
\fi
\end{document}